\documentclass[letterpaper]{article} % DO NOT CHANGE THIS
\usepackage{aaai2026}  % DO NOT CHANGE THIS
\usepackage{times}  % DO NOT CHANGE THIS
\usepackage{helvet}  % DO NOT CHANGE THIS
\usepackage{courier}  % DO NOT CHANGE THIS
\usepackage[hyphens]{url}  % DO NOT CHANGE THIS
\usepackage{graphicx} % DO NOT CHANGE THIS
\usepackage{natbib}  % DO NOT CHANGE THIS AND DO NOT ADD ANY OPTIONS TO IT
\usepackage{caption} % DO NOT CHANGE THIS AND DO NOT ADD ANY OPTIONS TO IT
\usepackage{algorithm}
\usepackage{algorithmic}
\usepackage{bm}
\usepackage{multirow}
\usepackage{amsmath}
\usepackage{amssymb}
\usepackage{booktabs}
\usepackage{newfloat}
\usepackage{listings}
\DeclareCaptionStyle{ruled}{labelfont=normalfont,labelsep=colon,strut=off} % DO NOT CHANGE THIS
\floatstyle{ruled}
\newfloat{listing}{tb}{lst}{}
\floatname{listing}{Listing}
\title{Beyond Fully Supervised Pixel Annotations: Scribble-Driven Weakly-Supervised Framework for Image Manipulation Localization}
\author{
    Songlin Li\textsuperscript{\rm 1},
    Guofeng Yu\textsuperscript{\rm 1},
    Zhiqing Guo\textsuperscript{\rm 1,2}\thanks{Corresponding author: Zhiqing Guo, guozhiqing@xju.edu.cn},
    Yunfeng Diao\textsuperscript{\rm 3},
    Dan Ma\textsuperscript{\rm 1},
    Gaobo Yang\textsuperscript{\rm 4}
}
\affiliations{
    \textsuperscript{\rm 1}School of Computer Science and Technology, Xinjiang University\\
    \textsuperscript{\rm 2}Xinjiang Multimodal Intelligent Processing and Information Security Engineering Technology Research Center\\
    \textsuperscript{\rm 3}School of Computer Science and Information Engineering, Hefei University of Technology\\
    \textsuperscript{\rm 4}College of Computer Science and Electronic Engineering, Hunan University\\
}

\usepackage{bibentry}
\begin{document}

\maketitle

\begin{abstract}
Deep learning-based image manipulation localization (IML) methods have achieved remarkable performance in recent years, but typically rely on large-scale pixel-level annotated datasets. To address the challenge of acquiring high-quality annotations, some recent weakly supervised methods utilize image-level labels to segment manipulated regions. However, the performance is still limited due to insufficient supervision signals. In this study, we explore a form of weak supervision that improves the annotation efficiency and detection performance, namely scribble annotation supervision. We re-annotate mainstream IML datasets with scribble labels and propose the first scribble-based IML (Sc-IML) dataset. Additionally, we propose the first scribble-based weakly supervised IML framework. Specifically,  we employ self-supervised training with a structural consistency loss to encourage the model to produce consistent predictions under multi-scale and augmented inputs. In addition, we propose a prior-aware feature modulation module (PFMM) that adaptively integrates prior information from both manipulated and authentic regions for dynamic feature adjustment, further enhancing feature discriminability and prediction consistency in complex scenes. We also propose a gated adaptive fusion module (GAFM) that utilizes gating mechanisms to regulate information flow during feature fusion, guiding the model toward emphasizing potential manipulated regions. Finally, we propose a confidence-aware entropy minimization loss (${\mathcal{L}}_{ {CEM }}$). This loss dynamically regularizes predictions in weakly annotated or unlabeled regions based on model uncertainty, effectively suppressing unreliable predictions. Experimental results show that our method outperforms existing fully supervised approaches in terms of average performance both in-distribution and out-of-distribution.
\end{abstract}

\begin{links}
    \link{Code}{https://github.com/vpsg-research/SCAF}
\end{links}

\section{Introduction}
Maliciously manipulated images can spread false information and seriously threaten social harmony. To safeguard information security and uncover the truth behind image manipulation, image manipulation localization (IML) technology, which focuses on accurately segmenting the manipulated regions, has been extensively studied.

\begin{figure}[!t]
  \centering
  \includegraphics[width=\linewidth]{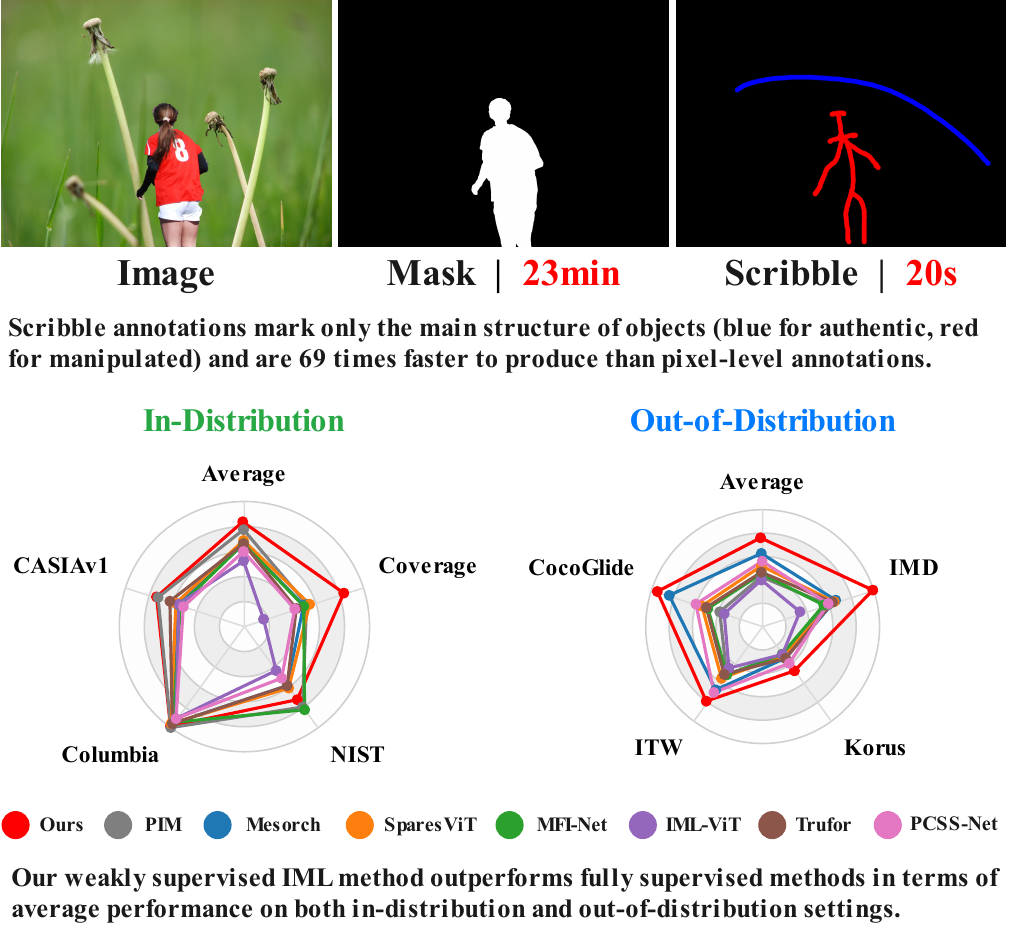}
  \caption{Comparative results between our model and existing IML methods in labeling efficiency and performance.}
  \label{fig_1}
\end{figure}

In recent years, deep learning techniques have significantly advanced the performance of IML methods, particularly when supported by large-scale pixel-level annotated datasets. Fully supervised methods can leverage dense pixel annotations to learn fine-grained manipulation features, achieving state-of-the-art localization results on various benchmark datasets. However, acquiring high-quality pixel-level annotations is both time-consuming and labor-intensive, which greatly limits the scalability and practical application of these methods. In real-world scenarios, obtaining large-scale high-precision pixel annotations is often infeasible, especially as data diversity and scale continue to expand. To reduce annotation costs, weakly supervised IML methods have emerged in recent years, typically relying on image-level labels to guide the localization of manipulated regions. While such methods alleviate the dependence on dense annotations, their limited supervision signals lead to clear bottlenecks in localization accuracy and generalization capability. A substantial performance gap still exists between weakly supervised and fully supervised methods.

To address these challenges, this paper explores a form of weak supervision, namely scribble annotations, which takes into account the efficiency of labeling and the amount of supervision information, as shown in Fig.~\ref{fig_1}. Scribble annotations enable rapid labeling of large-scale IML datasets while providing informative cues for localization tasks. We re-annotated several mainstream IML datasets, including 5,123 images from CASIAv2~\cite{dong2013casia}, 70 images from Coverage~\cite{wen2016coverage}, 130 images from Columbia~\cite{hsu2006columbia}, and 414 images from NIST16~\cite{guan2019mfc}, resulting in a total of 5,737 images. This constitutes the first scribble-based IML (Sc-IML) dataset. During annotation, annotators used CVAT to draw scribbles on the manipulated regions based on their first impression, without referring to the ground truth. To ensure high-quality data, the annotation process was cross-verified by three reviewers. Each image took approximately 20 seconds to annotate. However, existing studies have not reported the time required to manually annotate a pixel-level mask for a manipulated image. To address this issue, we organized 10 experienced computer vision researchers. Each researcher randomly selected 10 images from the training dataset for pixel-level annotation, with an average annotation time of approximately 23 minutes per image. This shows that scribble annotation is 69 times faster than pixel-level annotation. Although scribble annotations are significantly more efficient and convenient compared to pixel-level masks, they still face two major limitations: (1) Scribble annotation is a highly subjective form of weak supervision. Different annotators may have varying interpretations of the manipulated regions, boundaries, and details, resulting in significant inconsistency among the scribble annotations. (2) Scribble labels offer limited pixel-level supervision, causing the model to lack confidence in classifying unmarked regions and leading to prediction uncertainty.

Based on this, we propose the first weakly supervised IML framework utilizing scribble annotations, namely SCAF. To address the challenges of prediction inconsistency and uncertainty under scribble-based weak supervision in IML datasets, we introduce a series of innovative strategies. To resolve inconsistency, we employ self-supervised training with a structural consistency loss, which constrains the model to produce consistent predictions under multi-scale and augmented inputs. We also propose a prior-aware feature modulation module (PFMM) that adaptively integrates prior information from both manipulated and authentic regions for dynamic feature adjustment, and incorporate coordinate attention to efficiently model spatial dependencies, thereby significantly enhancing the discriminability and scene adaptability of feature representations. To tackle prediction uncertainty, we propose a gated adaptive fusion module (GAFM) that regulates information flow through multi-branch channel splitting and adaptive dynamic fusion. This module not only highlights critical features and suppresses redundancy, but also leverages scribble supervision to guide the model’s attention toward potential manipulated regions. Additionally, we propose a confidence-aware entropy minimization loss (${\mathcal{L}}_{ {CEM}}$), which dynamically filters model outputs based on uncertainty and applies adaptive entropy regularization to weakly annotated or unlabeled regions. This effectively suppresses unreliable predictions and improves model confidence and generalization in key areas. In summary, our contributions to IML are as follows:

\begin{itemize}

\item We propose the Sc-IML, the first scribble annotated dataset specifically designed for weakly supervised IML. Sc-IML effectively bridges the gap between costly pixel-level annotations and coarse image-level supervision by providing valuable spatial cues for the development and evaluation of weakly supervised IML methods, thereby advancing research in this field.

\item We propose the first weakly supervised IML framework based on scribble annotations, which outperforms existing fully supervised methods in terms of both in-distribution and out-of-distribution average performance.

\item We propose a PFMM that adaptively integrates prior information from both manipulated and authentic regions to achieve dynamic feature modulation. We further incorporate coordinate attention to efficiently model spatial dependencies, thereby significantly enhancing feature discrimination and scene adaptability.

\item We propose a GAFM that regulates information flow through multi-branch channel splitting and adaptive dynamic fusion. This module not only highlights critical features and suppresses redundancy, but also leverages scribble annotations to guide the model’s attention toward potential manipulated regions.

\item We propose a confidence-aware entropy minimization loss, which adaptively regularizes predictions in weakly annotated or unlabeled regions, significantly suppresses unreliable predictions, and enhances model confidence and generalization in critical areas.
\end{itemize}

\begin{figure*}[!t]
\centering
\includegraphics[width=\linewidth]{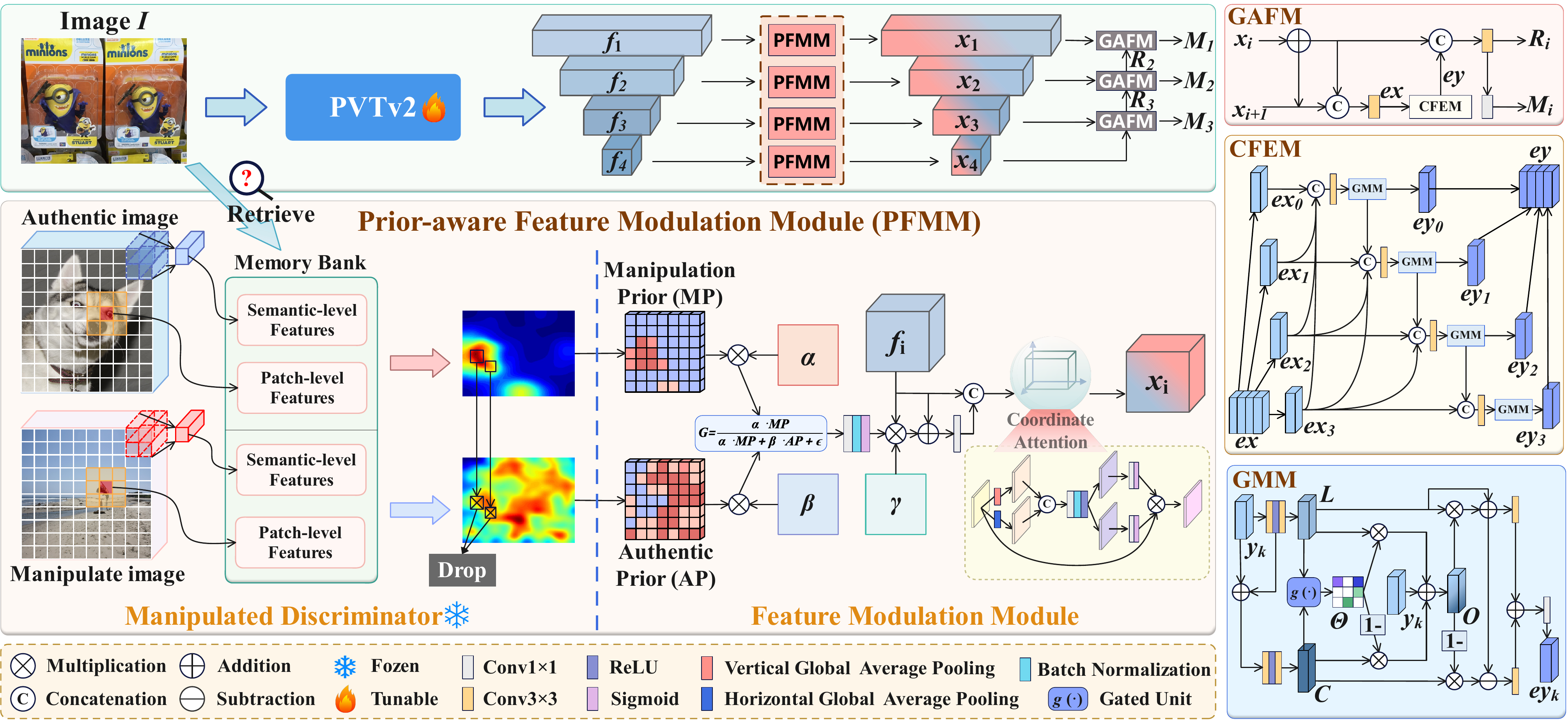}
% \caption{The overall architecture of the proposed method. The upper part of figure shows the training process, while the lower part presents the structure of SCAF. During training, we apply random transformations \( T(\cdot) \) to the input image. Both the original image \( I \) and the transformed image \( T(I) \) are fed into the network. For \( M_1 \), we employ the context affinity loss (\( \mathcal{L}_{CA} \)) and uncertainty-filtered entropy loss (${\mathcal{L}}_{ {UE}}$). Furthermore, \( M_1 \) is supervised with a structural consistency loss (\( \mathcal{L}_{SC} \)) against \( M_1' \), and a pixel-wise cross-entropy loss (\( \mathcal{L}_{PCE} \)) using the scribble annotation. For \( M_2 \) and \( M_3 \), only \( \mathcal{L}_{CA} \) and \( \mathcal{L}_{PCE} \) are adopted. The SCAF utilizes PVTv2 \cite{wang2022pvt} as the backbone and incorporates three main modules, namely the manipulated discriminator, the  feature modulation module (FMM), and the gated adaptive fusion module (GAFM).}
\caption{The overall architecture of the proposed SCAF. The model comprises two key modules: the prior-aware feature modulation module (PFMM) and the gated adaptive fusion module (GAFM). It is worth noting that the PFMM consists of a manipulated discriminator (MD) and a feature modulation module (FMM).}
\label{fig:NFF-Net}
\end{figure*}

% \cite{guillaro2023trufor} proposed an IML framework that combines RGB data with noise-resistant fingerprints for IML.
% \cite{guillaro2023trufor} proposed an IML framework that fuses RGB information with noise-sensitive fingerprints, enabling generalized detection of various image manipulations and achieving precise localization of tampered regions through anomaly detection and reliability assessment.
\section{Related Work}
\textbf{Fully supervised image manipulation localization: }
With the continuous development of deep learning and the emergence of large-scale datasets, fully supervised IML methods have achieved remarkable results.  For example, \cite{guillaro2023trufor} proposed an IML framework that combines RGB data with noise-resistant fingerprints for IML.  \cite{chen2024ean} effectively enhances the detection of boundary artifacts in IML by refining edge features and leveraging multi-feature fusion. \cite{10.1504/ijaacs.2024.139383} employs a Siamese network to extract illumination features and integrate them with a U-shaped network to localize manipulated regions. \cite{10.1504/ijaacs.2024.142523} embeds manipulation cues into similarity features to guide the matching process toward suspicious regions. \cite{10883001} exploits pixel inconsistencies to model global and local dependencies, enabling robust generalization to diverse manipulations. 

\textbf{Weakly supervised image manipulation localization: }
To reduce annotation costs, weakly supervised IML methods have been developed to segment manipulated regions using only image-level labels during training. \cite{zhai2023towards} applies self-consistency learning with multi-source cues to improve localization. \cite{ZHOU2024123501} iteratively refines pseudolabels to sharpen boundaries. \cite{10889843} introduces a cross-contrastive multi-stream fusion network that dispenses with pixel-level annotations. ~\cite{bai2025weakly} proposes WSCCL, which exploits multiple point correspondences for IML. 

Although fully supervised methods achieve high localization accuracy, they rely on large-scale high-quality pixel-level annotations, which are costly and time-consuming to obtain. Their scalability is further limited by the lack of such data in real-world scenarios. Weakly supervised IML methods using image-level labels greatly reduce annotation costs but lack spatial cues, making precise localization difficult and often missing subtle manipulations. To address these challenges, we propose a novel weakly supervised IML approach with scribble annotations. Scribbles provide crucial spatial information to the model, guiding it to learn the spatial distribution of manipulated regions while significantly improving annotation efficiency. Our model outperforms existing fully supervised methods in terms of average performance, both in-distribution and out-of-distribution.

\begin{figure}[!t]
  \centering
  \includegraphics[width=\linewidth]{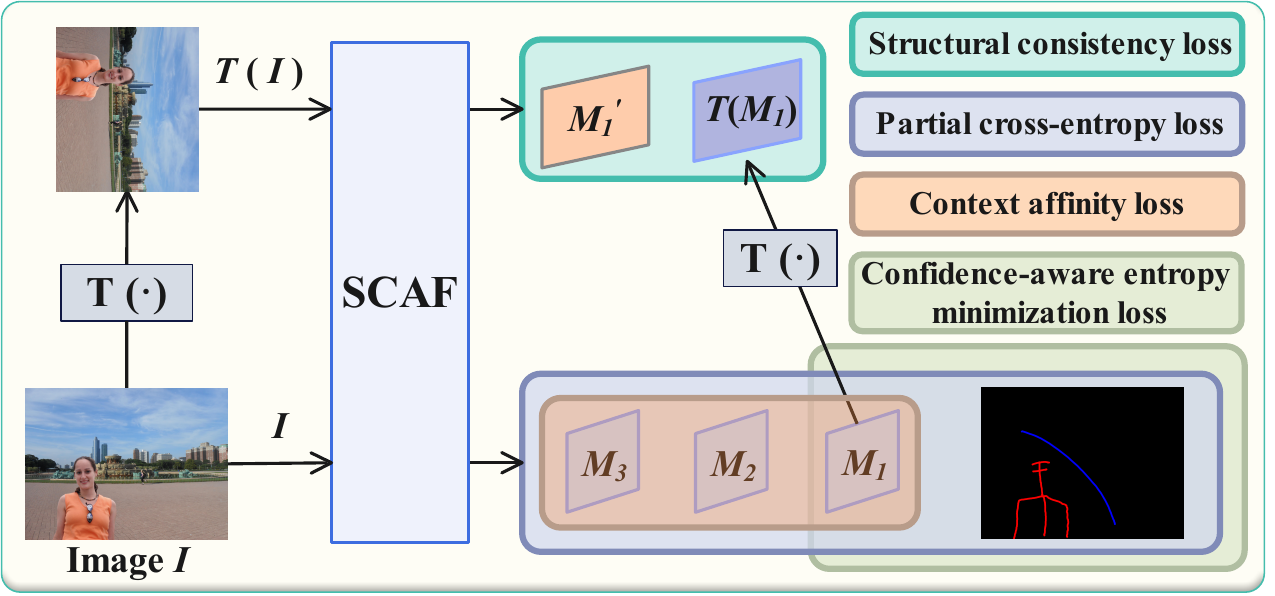}
  \caption{The training process of the proposed model. $\bm{T}(\cdot)$ denotes random rotations, scaling, and flipping.}
  \label{train}
\end{figure}

\section{Methodology}
\subsection{Overview}

The overall architecture of SCAF is shown in Fig.~\ref{fig:NFF-Net}. Specifically, the input image $\bm{I}$ is first processed by PVTv2~\cite{wang2022pvt} to extract multi-scale features $\bm{f}_i$, and these features are then modulated in the prior-aware feature modulation module (PFMM) using the prior to integrate them into $\bm{x}_i$. Finally, the features are fused by the gated adaptive fusion module (GAFM) to produce the final predicted mask. The training process of the model is illustrated in Fig.~\ref{train}. An image $\bm{I}$ is input into the SCAF to generate the primary output $\bm{M}_1$ as well as auxiliary outputs $\bm{M}_2$ and $\bm{M}_3$. Additionally, a randomly transformed version of $\bm{I}$, denoted as $T(\bm{I})$, which may include rotation, scaling, or flipping, is also fed into the SCAF to obtain $\bm{M}_1^{\prime}$. This, together with $\bm{M}_1$, is used to compute the structural consistency loss (${\mathcal{L}}_{{SC}}$)~\cite{he2023weakly}. Furthermore, $\bm{M}_1$ is jointly supervised by the partial cross-entropy loss (${\mathcal{L}}_{{PCE}}$), context affinity loss (${\mathcal{L}}_{{CA}}$)~\cite{obukhov2019gated}, and confidence-aware entropy minimization loss (${\mathcal{L}}_{{CEM}}$), while $\bm{M}_2$ and $\bm{M}_3$ are optimized using only ${\mathcal{L}}_{{CA}}$ and ${\mathcal{L}}_{{PCE}}$.

\subsection{Prior-aware Feature Modulation Module}
For scribble annotations, annotators typically mark the manipulated regions in an image by freely drawing rough scribbles. Since this annotation method heavily relies on personal subjective judgment, different annotators may have varying understandings of the manipulated regions, interpretations of boundaries, and levels of attention to detail. Even for the same image, the content of the scribble annotations can differ significantly. This subjectivity and randomness lead to considerable label inconsistency within the training set, which in turn affects the model’s ability to accurately identify manipulated regions and generalize to new data. In complex scenarios, this often results in unstable and unreliable predictions. To address the inconsistency caused by subjective scribble annotations, we propose a prior-aware feature modulation module (PFMM). As shown in the lower left of Fig.~\ref{fig:NFF-Net}, PFMM is composed of two parts: the manipulated discriminator (MD) and the feature modulation module (FMM). The core idea is to utilize manipulated priors (MP) and authentic prior (AP) generated by the MD, and then leverage these priors within the FMM to dynamically modulate the features $\bm{f}_i$ in a targeted manner. By explicitly introducing prior knowledge, PFMM can effectively mitigate the noise and bias introduced by subjective scribble annotations, guiding the model to focus on more objective region distributions. This mechanism enhances the model’s discriminability for manipulated regions and improves prediction consistency and robustness.

\textbf{1) Manipulated discriminator (MD):} Existing research~\cite{roth2022towards} has demonstrated that patch-level features are effective for detecting subtle anomalies. However, they lack explicit suppression of semantic bias, which becomes particularly evident in cross-domain scenarios. To address this limitation, we propose a selective suppression mechanism based on semantic-level features, which builds upon patch-level representations. By memorizing multi-scale semantics from both authentic and manipulated images, our method adaptively suppresses features that are highly similar to those stored in the memory bank during inference. This enables the generation of precise authentic region priors and manipulated region priors. Specifically, taking the training process of authentic images as an example, we first extract multi-layer features using the PVTv2. Each layer’s feature map is then partitioned into overlapping patches with a stride of 1, resulting in a set of local patch features ${\bm{p}_j^s}$. To further incorporate local contextual information, we perform weighted fusion of each patch and its eight spatial neighbors to obtain $\tilde{\bm{p}}_{j}$, as formulated below:
\begin{equation}
    \tilde{\bm{p}}_{j}^s=\sum_{n \in \mathcal{N}(j)} w_{n} \bm{p}_{n}^s
\end{equation}
where  $\mathcal{N}(j)$ denotes the $3 \times 3$ neighborhood centered at $j$, and $w_j$ represents the weighting coefficient. The collection of $\tilde{\bm{p}}_{j}^s$ constitutes the patch-level feature $\bm{P}_s$ for an image. We store the patch-level features $\bm{P}_s$ of all images in the training set to construct a patch-level memory bank $\mathcal{B}_p$:
\begin{equation}
 \left\{\begin{array}{l}
    \bm{P}_s =\left\{\tilde{\bm{p}}_{j}^s \mid j=1, \ldots, T\right\} \\
    \mathcal{B}_p =\left\{{\bm{P}}_{s} \mid s=1, \ldots, N\right\}
    \end{array}\right.
\end{equation}
where $T$ denotes the total number of local patch features in a sample. $N$ denotes the number of training samples. For the construction of the semantic-level memory bank $\mathcal{B}_s$, we first extract features from the authentic training samples using the backbone network, and reduce the channel dimension of the extracted features to 256 to obtain the feature set $\left\{\bm{v}_{m}\right\}_{m=1}^{N}$. These features are then subjected to L2 normalization. Finally, the semantic features of all training images are stored to form the memory bank:
\begin{equation}
        \left\{\begin{array}{l}
\tilde{\bm{v}}_{m}={\bm{v}_{m}} \:/ \:{\left\|\bm{v}_{m}\right\|_{2}}\\
\mathcal{B}_s =\left\{\tilde{\bm{v}}_{m} \mid m=1, \ldots, N\right\}
\end{array}\right.
\end{equation}

The model constructs a memory bank of authentic region features by learning from a large number of real images. However, this bank is filled with highly similar features, resulting in redundancy and bias, which leads the model to overgeneralize dominant authentic region features. During inference, this mechanism causes subtle manipulated features to be easily overwhelmed by the abundance of highly matching authentic features, thereby reducing the saliency of manipulated regions and ultimately lowering localization accuracy. To address this issue, we actively identify and suppress redundant authentic feature noise, which relatively amplifies and highlights manipulated features that represent inconsistencies. In this way, the model shifts from passively learning authentic image features to actively capturing the distinctive traces of manipulation. The details are as follows:
\begin{equation}
  \bm{q}_{sup} =\left( 1-\max _{m}\left(\frac{\bm{q} \times \bm{v}_{m}}{\left\|\bm{q}\right\|_2\left\|\bm{v}_{m}\right\|_2}\right)\right) \times \bm{q}
\end{equation}
where $\bm{q}$ denotes the current input feature, which is compared with each feature $\bm{v}_m$ in $\mathcal{B}_s$ to compute the maximum cosine similarity. This process suppresses regions that are overly similar to authentic features from the training set and highlights manipulated features, resulting in $\bm{q}_{{sup}}$. Finally, the Euclidean distance between $\bm{q}_{sup}$ and its nearest feature $\bm{P}_s$ in $\mathcal{B}_p$ is calculated. It can be formulated as follows:
\begin{equation}
    \operatorname{Score}(\bm{q}_{sup})=\left\|\bm{q}_{sup}-\bm{P}_{s}\right\|_{2}
\end{equation}

By scoring each location in the input image, we obtain a prior map of manipulated regions. Similarly, training on manipulated images yields a prior map for authentic regions. However, since manipulated images contain both authentic and manipulated areas, the authentic prior may include false activations. To address this, we compute the cosine similarity between the manipulated and authentic priors, removing falsely activated regions to obtain a purified authentic prior.

\textbf{2) Feature modulation module (FMM):} FFM is designed to adaptively modulate the feature $\bm{f}_i$ using prior information, thereby alleviating the adverse effects caused by the inconsistency of scribble annotations. Specifically, FFM utilizes learnable manipulated region weight $\alpha$ and authentic region weight $\beta$ to perform weighted normalization of the prior knowledge, resulting in the computation of the manipulated region probability response $\bm{G}$, as formulated below:
\begin{equation}
\bm{G} = \frac{\alpha \times MP}{\alpha \times MP + \beta \times AP + \epsilon}
\end{equation}
where $\epsilon$ is a small constant added to prevent division by zero. This controllable probabilistic modeling approach allows the module to adaptively adjust its sensitivity to different regions based on the data distribution, effectively reducing the impact of label inconsistency or prior noise on performance. Then, $\bm{G}$ is passed through a $1 \times 1$ convolution, batch normalization, and a sigmoid activation function to generate an enhanced feature map $\bm{Ge}$ for each spatial location. This is further regularized residually by a learnable parameter $\gamma$ to obtain the feature $\bm{F}$, as follows:
\begin{equation}
\bm{F} = \operatorname{Conv1}(\bm{f}_i + \gamma \times \bm{Ge} \times \bm{f_i})
\end{equation}
where $\operatorname{Conv1}$ denotes a $1 \times 1$ convolution. Finally, coordinate attention is applied to the concatenated features of $\bm{F}$ and $\bm{f}_i$, resulting in the output feature $\bm{x}_i$. Coordinate attention captures both spatial location information and channel interdependencies, further enhancing the model’s ability to localize and discriminate manipulated regions.

% \begin{equation}
%     \bm{x}_i  = CoordAtt(\operatorname{Conv1}(Cat(\bm{f}_i, \: \bm{F})))
% \end{equation}

In summary, PFMM explicitly incorporates manipulated and authentic priors to effectively alleviate the adverse effects of inconsistency and label noise introduced by scribble annotations. By leveraging prior-guided feature modulation, the model is able to focus on more objective and reliable regional cues, thereby enhancing both the accuracy and generalization ability of tampering localization.

\subsection{Gated Adaptive Fusion Module}
% In scribble-based weakly supervised IML, the annotations are often imprecise and lack sufficient information, which introduces uncertainty in model predictions. To address this issue, we propose a gated adaptive fusion module (GAFM) that substantially enriches contextual information and leverages scribble annotations to guide the model in expanding its attention to potential manipulated regions, thereby improving the model’s accuracy, as illustrated in Fig.~\ref{gafm}.
In scribble-based weakly supervised IML, annotations are sparse and irregular, making it challenging for models to fully localize manipulated regions. This incomplete supervision introduces significant uncertainty during training and hampers the detection of subtle manipulations, often leading to missed or false detections. Furthermore, manipulated regions can exhibit diverse spatial distributions and weak visual clues, which are hard to capture using conventional feature aggregation or single-scale modeling. To address these issues, we propose a gated adaptive fusion module (GAFM), as shown on the right side of Fig.~\ref{fig:NFF-Net}, which consists of a core channel-aware feature enhancement module (CFEM) and a gated modulation module (GMM). The GAFM aggregates multi-scale contextual features and employs both the CFEM and GMM to group channels and progressively perform gated fusion, thereby adaptively suppressing feature uncertainty and enhancing the richness and discriminability of the feature representations. Specifically, the fused feature $\bm{e}_x$, obtained by merging $\bm{x}_i$ and $\bm{x}_{i+1}$, is first input into CFEM, where it is evenly split into four groups along the channel dimension.
\begin{equation}
    \bm{ex} = [\bm{ex}_0,\, \bm{ex}_1,\, \bm{ex}_2,\, \bm{ex}_3], \quad
    \bm{ex}_k \in \mathbb{R}^{B \times \frac{C}{4} \times H \times W}
\end{equation}
Then, each group feature $\bm{ex}_j$ is concatenated with its neighboring groups or the output of the previous GMM, and sequentially passed through a $3 \times 3$ convolution and the GMM, forming a progressive information fusion flow.
\begin{equation}
    \left\{\begin{array}{l}
\bm{y}_{k}=\operatorname{Conv3}( Cat({ Cat }\left.\right|_{z=k} ^{3} \bm{ex}_{z},\: \mathbb{I}(k>0) \times \bm{ey}_{k-1})) \\
\bm{ey}_{k}=\operatorname{GMM}\left(\bm{y}_{k}\right)
\end{array}\right.
\end{equation}
where $\operatorname{Conv3}$ denotes a $3 \times 3$ convolution. ${ Cat }\left.\right|_{z=k} ^{3}$ denotes the concatenation of all $\bm{ex}_n$ for $n=j$ to 3 in sequence. $\mathbb{I}(k>0)$ denotes the indicator function, which returns 1 if $k>0$, and 0 otherwise. The GMM employs a gating mechanism to achieve adaptive recalibration and interaction between local and global features, thereby enhancing the modeling capability for fine-grained representations. Specifically, the feature $\bm{y}_k$ is first processed through two separate sequences of $3 \times 3$ convolutional layers followed by ReLU activations, resulting in two distinct feature branches.
\begin{equation}
    \left\{\begin{array}{l}
\bm{L} = \operatorname{Conv3}(ReLU(\operatorname{Conv3}(\bm{y}_k))) \\
\bm{C} = \operatorname{Conv3}(ReLU(\operatorname{Conv3}(\bm{L} + \bm{y}_k)))
\end{array}\right.
\end{equation}

% \begin{figure}[!t]
%   \centering
%   \includegraphics[width=\linewidth]{CameraReady/LaTeX/figure3.pdf}
%   \caption{The upper part illustrates the architecture of the channel-aware feature enhancement module (CFEM), while the lower right part presents the structure of the core component, the gated modulation module (GMM), within CFEM.}
%   \label{gafm}
% \end{figure}

Subsequently, the extracted local-detail feature $\bm{L}$ and contextual semantic feature $\bm{C}$ are concatenated along the channel dimension and fed into a gating unit $g(\cdot)$~\cite{10038722}. By effectively leveraging the complementary relationship between these two types of features, the gating unit adaptively generates a more accurate gating coefficient $\alpha$, enabling precise modulation and enhancement of features of the manipulated region. It can be formulated as:
\begin{equation}
\left\{\begin{array}{l}
    \theta = g(Cat(\bm{L}, \: \bm{C})) \\
    \bm{O} = \bm{y}_j + \theta \times \bm{L}+(1-\theta) \times \bm{C}
    \end{array}\right.
\end{equation}

Due to the limited pixel-level supervision provided by scribble-based annotations, the model may confuse features of manipulated regions with those of the authentic background, leading to ambiguous localization. To address this, we construct a complementary reverse mask ($1-\bm{O}$) from the feature map $ \bm{O}$ and combine it with feature $\bm{C}$ to obtain the authentic background features. We enhance the manipulated region’s feature representation $\bm{O}$ by combining it with feature $\bm{L}$ and applying a residual connection. Subsequently, the difference between the manipulated-enhanced feature $\bm{Me}$  and the authentic background features is computed to highlight subtle or latent anomalies within the manipulated regions, resulting in the differential feature $\bm{D}$. Then, the manipulated-enhanced feature $\bm{Me}$ is fused with the differential feature $\bm{D}$ to obtain the output feature $\bm{ey}_j$, which can be expressed as $\bm{ey}_k = \operatorname{Conv1}(\bm{Me}+\bm{D})$.

% \begin{equation}
% \left\{\begin{array}{l}
%     \bm{Me} =\operatorname{Conv3}(\bm{L} \times \bm{O} + \bm{L})  \\
%     \bm{A} = \operatorname{Conv3}(\bm{\bm{C}}\times(1-\bm{O})) \\
%     \bm{D} = \operatorname{Conv3}(\bm{Me}-\bm{A})
%     \end{array}\right.
% \end{equation}

The GMM enables the model to adaptively focus on manipulated regions, enhancing manipulated features while suppressing background features. Finally, we concatenate the outputs from multiple stages of the GMM to obtain the feature $\bm{e}y$, which can be expressed as $\bm{ey} = { Cat }\left.\right|_{k=0} ^{3} \bm{ey}_k$.
% \begin{equation}
%     \bm{ey} = { Cat }\left.\right|_{k=0} ^{3} \bm{ey}_k
% \end{equation}

Since each stage of the GMM focuses on adaptive modeling of specific sub-channel features, these features are inherently complementary. By concatenating them along the channel dimension, the model maximally preserves the heterogeneous information across groups, thereby providing richer contextual cues for subsequent integration and enhancing the perception and representation of complex manipulated regions.

\subsection{Confidence-aware Entropy Minimization Loss}
In scribble-based weakly supervised IML, the sparsity of supervision signals makes it difficult to adequately constrain the model’s learning. While entropy minimization is an effective strategy for leveraging unlabeled regions, blindly applying it may cause the model to become “over-confident” on unreliable predictions, amplifying errors and resulting in unstable training. To address this issue, we propose a confidence-aware entropy minimization loss (${\mathcal{L}}_{ {CEM}}$). This loss applies entropy minimization only to unlabeled pixels whose current predictions are deemed reliable based on a confidence threshold. Additionally, a weak regularization term is applied to the annotated regions to directly encourage the model to produce predictions with higher confidence and sharper boundaries. Specifically, the model’s primary output is $\bm{M}_1 \in \mathbb{R}^{B \times 1 \times H \times W}$, where each pixel is associated with a probability value $m_t \in [0,1]$. The Shannon entropy $\mathcal{H}(\cdot)$ for each pixel is defined as follows:
\begin{equation}
    \mathcal{H}\left(\bm{m}_{t}\right)=-\left[\bm{m}_{t} \log \bm{m}_{t}+\left(1-\bm{m}_{t}\right) \log \left(1-\bm{m}_{t}\right)\right]
\end{equation}

To avoid introducing noise from regions where the model’s predictions are unreliable, we perform entropy minimization only on unlabeled pixels where the model already exhibits high confidence. The loss $\mathcal{L}_{un}$ is defined as the average entropy over unlabeled pixels whose predicted entropy is below a threshold of 0.5:
\begin{equation}
    \mathcal{L}_{ {un}}=\frac{\sum_{t \in \mathcal{U}} \mathcal{H}\left(\bm{m}_{t}\right) \cdot \mathbb{I}\left(\mathcal{H}\left(\bm{m}_{t}\right)<0.5\right)}{\sum_{t \in \mathcal{U}} \mathbb{I}\left(\mathcal{H}\left(\bm{m}_{t}\right)<0.5\right)+\epsilon}
\end{equation}
where $\mathcal{U}$ denotes the set of unlabeled pixels. Meanwhile, we also compute the average entropy over labeled pixels as a weak regularization penalty, encouraging the model to make confident predictions in known regions. This loss, denoted as $\mathcal{L}_{ {la}}$, is defined as follows:
\begin{equation}
    \mathcal{L}_{ {la }}=w_{weak}\cdot\frac{\sum_{t \in \mathcal{G}} \mathcal{H}\left(\bm{m}_{t}\right)}{|\mathcal{G}|+\epsilon}
\end{equation}
where $\mathcal{G}$ denotes the set of labeled pixels and $|\mathcal{G}|$ is the total number of labeled pixels. The final mixed entropy loss $\mathcal{L}_{ent}$ is defined as the sum of $\mathcal{L}_{un}$ and $\mathcal{L}_{la}$. To ensure stable training, we adopt a weight ramp-up strategy, introducing a dynamic weight $\lambda(T)$ to modulate $\mathcal{L}_{ent}$.
\begin{equation}
    \lambda(T) = w_{\text {max }} \cdot \exp \left(-\left(1-\frac{\min \left(T, { {ramp }}\right)}{{ {ramp }}}\right)^{2}\right)
\end{equation}
where $T$ denotes the current epoch and $ramp$ represents the ramp-up period. During this period, $\lambda(T)$ increases smoothly from 0 to its maximum value $w_{max}$. In this work, we set $w_{max}$ and $w_{weak}$ to 0.1, which is chosen to balance training stability and final model performance. Thus, the final loss term is defined as $\mathcal{L}_{CEM} = \lambda(T) \cdot\mathcal{L}_{ent}$.

\begin{table}[!t]
\centering

 \small % 这是唯一允许的字体大小调整命令
 \setlength{\tabcolsep}{1mm}

% \resizebox{0.47\textwidth}{!}{%
\begin{tabular}{cccccccc}
\toprule
Method  & CASIAv1 & Coverage & NIST & IMD & Columbia  \\
\midrule
WSCL  & 0.153 & 0.201 & 0.099 & 0.173 & 0.362  \\
EdgeCAM  &0.301 & 0.262 & 0.254 & 0.242 & 0.470  \\
SOWCL   &0.334 & 0.239 & 0.288 & 0.259 & 0.385  \\
WSCCL  & 0.349 & 0.281 & 0.278 & 0.259 & 0.516  \\
Ours & \textbf{0.716} & \textbf{0.827} & \textbf{0.721} & \textbf{0.580} & \textbf{0.979} \\
\bottomrule
\end{tabular}
% }
\caption{Comparison with other advanced weakly supervised IML methods. }
\label{tab:comparison_transposed}
\end{table}

\section{Experiments and Results}
\subsection{Datasets and Implementation Details}
Our experiments primarily utilize 8 mainstream benchmark datasets: NIST~\cite{guan2019mfc},  CASIA~\cite{dong2013casia}, Columbia~\cite{hsu2006columbia}, Coverage~\cite{wen2016coverage}, IMD~\cite{Novozamsky_2020_WACV}, CocoGlide~\cite{nichol2021glide}, ITW~\cite{huh2018fighting}, and Korus~\cite{korus2016evaluation}. We adopt the same training set split as \cite{zhou2024contribution}. During training, all images are resized to $512 \times 512$. The model is trained with a batch size of 32 using the AdamW optimizer. The initial learning rate is set to 1e-4 and decayed by a factor of 0.1 every 50 epochs. Training is performed on 4 NVIDIA 4090 GPUs with a total of 70 epochs.

\subsection{Comparison with SOTA Methods}
Pixel-level \textbf{F1} and \textbf{AUC} are standard metrics for IML, but recent research~\cite{ma2025imdl} shows that \textbf{AUC} exhibits overconfidence in IML. Therefore, we evaluate all experiments using the \textbf{F1-score} with a fixed threshold of 0.5.

\textbf{Image manipulation localization.} Table~\ref{tab:comparison_transposed} presents a comparison of all published weakly supervised IML methods to date, including WSCL~\cite{zhai2023towards}, EdgeCAM~\cite{ZHOU2024123501}, SOWCL~\cite{10889843}, and WSCCL~\cite{bai2025weakly}. While these methods rely on image-level labels, the results clearly demonstrate that our approach significantly outperforms them across all datasets, highlighting that the spatial constraints provided by scribble annotations lead to more accurate  localization. Table~\ref{sota} presents a comparison with several SOTA fully supervised methods, such as  PCSS-Net~\cite{liu2022pscc}, Trufor~\cite{guillaro2023trufor}, MFI-Net~\cite{ren2023mfi}, SparseViT~\cite{su2025can}, PIM~\cite{kong2025pixel}, and Mesorch~\cite{zhu2025mesoscopic}, evaluated under both in-distribution and out-of-distribution settings. Our method not only achieves higher average performance than fully supervised baselines under standard conditions, but also demonstrates stronger generalization to previously unseen manipulation scenarios. This enhanced generalization is attributed to the balance between annotation efficiency and effective supervision achieved by scribble annotations, which provide direct spatial guidance with minimal labeling effort and help prevent overfitting to dense pixel-level labels. It is worth noting that all fully supervised methods were retrained on the same benchmarks as ours for a fair comparison. However, since among the weakly supervised methods only WSCL has publicly available code, we retrained only WSCL, while the results for the other methods are taken directly from their published papers.
 % Furthermore, our framework integrates self-supervised structural consistency, prior-aware feature modulation, gated adaptive fusion, and confidence-aware entropy minimization, which collectively promote consistency, uncertainty awareness, and dynamic feature adjustment. These innovations enable our model to achieve higher accuracy and adaptability in both conventional and real-world settings.

 \begin{table*}[!t]
\centering
% \textbf{Average} denotes the average results across the four datasets.
 \small % 这是唯一允许的字体大小调整命令
 % \setlength{\tabcolsep}{1mm}

% \resizebox{\textwidth}{!}{%
\begin{tabular}{ccccccccccccc}
\toprule
\multirow{2}{*}{\textbf{Method}} & \multirow{2}{*}{\textbf{Pub.}} & \multicolumn{5}{c}{\textbf{In-Distribution (ID)}} & \multicolumn{6}{c}{\textbf{Out-of-Distribution (OOD)}}\\ \cmidrule(lr){3-7} \cmidrule(lr){9-13}
               & & {{NIST}} &{{CASIAv1}} & {{Columbia}} & {{Coverage}} & {{Avg.}} & & {{CocoGlide}} &{{ITW}} & {{Korus}}  & {{IMD}} & {{Avg.}} \\ \toprule
 % MVSS-Net  & ICCV'21 & 0.428 & 0.385 & 0.835 & 0.344 & 0.498 && 0.319 & 0.277 & 0.098  & 0.249 & 0.237 \\
PCSS-Net  & TCSVT'22 & 0.509 & 0.503 & 0.905 & 0.424 & 0.585 && 0.344 & \underline{0.407} & \underline{0.229}  & 0.350 & 0.333 \\

Trufor  & CVPR'23 & 0.584 & 0.603 & 0.953 & 0.435 & 0.644 && 0.287 & 0.310 & 0.200  & 0.375 & 0.293 \\
IML-ViT  & Arxiv'24 & 0.440 & 0.529 & 0.906 & 0.168 & 0.511 && 0.200 & 0.270 & 0.181  & 0.209 & 0.215 \\
MFI-Net  & TSCVT'24 & \textbf{0.817} & 0.524 & 0.938 & 0.497 & 0.644 && 0.283 & 0.300 & 0.186  & 0.329 & 0.275 \\
SparseViT  & AAAI'25 & 0.616 & 0.557 & 0.958 & \underline{0.546} & 0.669 && 0.311 & 0.333 & 0.193  & 0.381 & 0.305 \\
PIM                     &TPAMI'25 & 0.587 & 0.548 & 0.954 & 0.504 & 0.648 && \underline{0.481} & {0.392} & {0.203}  & \underline{0.395} & \underline{0.368}\\ 
Mesorch                       &AAAI'25 & \underline{0.802} & \underline{0.703} & \textbf{0.981} & {0.531} & \underline{0.754} && 0.218 & 0.299 & 0.182  & 0.346 & 0.261\\ 
Ours                           & -     &  {0.721} & \textbf{0.716} & \underline{0.979} & \textbf{0.827} & \textbf{0.811} && \textbf{0.546} & \textbf{0.467} & \textbf{0.282}  & \textbf{0.580} & \textbf{0.469}\\ \bottomrule
\end{tabular}%
% }
\caption{Comparison with other  fully supervised IML methods. Bold and underlined indicate the best and second-best.}
\label{sota}
\end{table*}

\begin{figure}[!t]
  \centering
  \includegraphics[width=\linewidth]{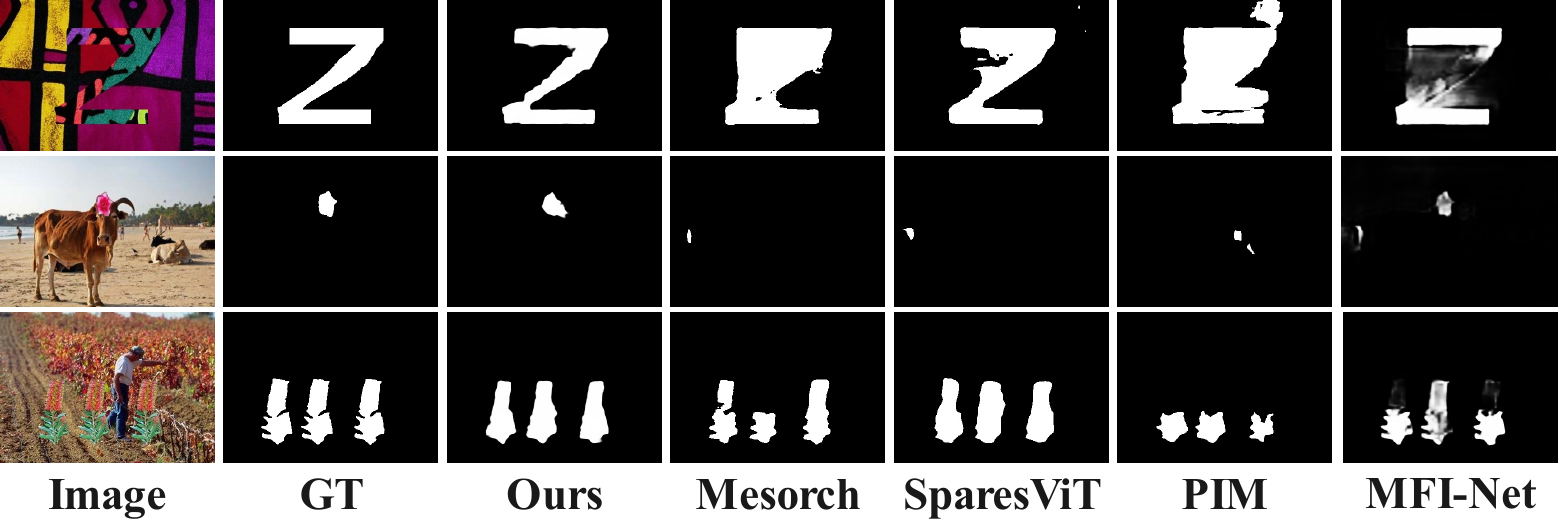}
\caption{Visualization results of different methods. }
  \label{fig:vs}
\end{figure}

\textbf{Visual comparison.} Fig.\ref{fig:vs} presents the segmentation results of our proposed model and several fully supervised methods under challenging scenarios, including large-scale manipulations, small-scale manipulations, and multi-object manipulations. Clearly, our method achieves the best visual performance. In large-scale manipulations, our model accurately identifies the manipulated regions, producing clear and complete segmentation, while the compared methods often suffer from false positives. In small-scale manipulations, our method provides precise localization, whereas the other methods almost entirely fail. For multi-object manipulations, our method not only detects all manipulated regions but also excels in capturing fine details.

\begin{table}[!t]
\centering

% \resizebox{0.47\textwidth}{!}{%
 \small % 这是唯一允许的字体大小调整命令

\begin{tabular}{ccccc}
\toprule 
Baseline & GAFM & PFMM  &  { \textbf{Avg.ID} } & { \textbf{Avg.OOD} }
 \\
\midrule
 $ \checkmark $  & & &  0.533&  0.283   \\
$ \checkmark $ & $ \checkmark $ & & 0.697 & 0.434  \\
 $ \checkmark $  &  $ \checkmark $   & $ \checkmark $  &\textbf{0.811} &  \textbf{0.469}  \\

\bottomrule
\end{tabular}
% }
\caption{The ablation study for our modules. }
\label{ab}
\end{table}

\begin{table}[!t]
\centering
% \resizebox{0.47\textwidth}{!}{%
 \small % 这是唯一允许的字体大小调整命令

\begin{tabular}{cccccc}
\toprule 
${\mathcal{L}}_{ {PCE}}$ & ${\mathcal{L}}_{ {CA}}$ & ${\mathcal{L}}_{ {SC}}$  & ${\mathcal{L}}_{ {CEM}}$ & { \textbf{Avg.ID} } & { \textbf{Avg.OOD} }
 \\
\midrule
 $ \checkmark $  & & &&  0.627 & 0.394  \\
$ \checkmark $ & $ \checkmark $ & & &0.731& 0.438  \\
 $ \checkmark $  &  $ \checkmark $   & $ \checkmark $  & &  0.755 & 0.414  \\
  $ \checkmark $  &  $ \checkmark $   &   & $ \checkmark $ &  0.733 & 0.418  \\
  $ \checkmark $  & $ \checkmark $  & $ \checkmark $  & $ \checkmark $ &\textbf{0.811} & \textbf{0.469} \\

\bottomrule
\end{tabular}
% }
\caption{The ablation study for our loss functions. }
\label{ab_loss}
\end{table}

% \begin{table}[!t]
% \centering
% \caption{Ablation study results of model components.}
% \label{ab}
%  \small % 这是唯一允许的字体大小调整命令
%  \setlength{\tabcolsep}{1mm}
% % \resizebox{0.47\textwidth}{!}{%
% \begin{tabular}{lccccc}
% \toprule
% {\textbf{Method}} & {\textbf{NIST}} & {\textbf{CASIAv1}} & {\textbf{Columbia}} & {\textbf{Coverage}} & {\textbf{Avg.}} \\ 

% \midrule
% (a) B& 0.404 & 0.396 & 0.841 & 0.621 & 0.566\\

% (b) B+GAFM& 0.668 & 0.540 & 0.960 & 0.618 & 0.697 \\ 

% (d) Ours &\pmb{0.721} & \pmb{0.716} & \pmb{0.979} & \pmb{0.827} & \pmb{0.811} \\ 
% \bottomrule
% \end{tabular}
% % }
% \end{table}

% \begin{table}[!t]
% \centering
% \caption{The ablation study for our loss functions. }
% \label{ab_loss}
% % \resizebox{0.47\textwidth}{!}{%
%  \small % 这是唯一允许的字体大小调整命令
%  \setlength{\tabcolsep}{1mm}
% \begin{tabular}{cccc|cccc}
% \toprule 
% ${\mathcal{L}}_{ {PCE}}$ & ${\mathcal{L}}_{ {CA}}$ & ${\mathcal{L}}_{ {SC}}$  & ${\mathcal{L}}_{ {CEM}}$ & { \textbf{NIST} } & { \textbf{CASIAv1} } & { \textbf{Columbia} } & { \textbf{Coverage} } 
%  \\
% \midrule
%  $ \checkmark $  & & &&  0.308 & 0.603 & 0.900 & 0.698   \\
% $ \checkmark $ & $ \checkmark $ & & & 0.642 & 0.654 & 0.933 & 0.694  \\
%  $ \checkmark $  &  $ \checkmark $   & $ \checkmark $  & &  0.618 & 0.707 & 0.962 & 0.732  \\
%   $ \checkmark $  & $ \checkmark $  & $ \checkmark $  & $ \checkmark $ &\pmb{0.721} & \pmb{0.716} & \pmb{0.979} & \pmb{0.827}   \\

% \bottomrule
% \end{tabular}
% % }
% \end{table}

\subsection{Ablation Study}
As shown in Table~\ref{ab}, introducing GAFM significantly improves performance, highlighting the importance of adaptive information regulation in feature fusion. When all modules are integrated, our model achieves the best results on both in-distribution (ID) and out-of-distribution (OOD) datasets, underscoring the synergistic effect of the proposed structural designs and feature modulation strategies. Table~\ref{ab_loss} further analyzes the contribution of each loss function. Introducing ${\mathcal{L}}_{CA}$ improves performance, confirming the benefit of context-aware modeling. While ${\mathcal{L}}_{SC}$ boosts ID results, it reduces flexibility on OOD datasets and increases prediction uncertainty, leading to performance degradation. ${\mathcal{L}}_{CEM}$ alone yields only marginal gains. When ${\mathcal{L}}_{SC}$ and ${\mathcal{L}}_{CEM}$ are combined, the model achieves substantial improvements on both ID and OOD datasets, indicating that structural consistency and confidence-aware entropy minimization are complementary: ${\mathcal{L}}_{SC}$ provides a stable structural prior, whereas ${\mathcal{L}}_{CEM}$ adaptively suppresses unreliable predictions in weakly annotated or unlabeled regions, thereby enhancing robustness and generalization.

\subsection{Robustness and Efficiency of the Model} 
With the rapid development of the internet, online social platforms have become a primary channel for image dissemination. To evaluate the robustness of our model under such conditions, we followed the same benchmark as~\citep{MVSS_2022TPAMI} and applied compression through platforms such as Facebook, Weibo, WeChat, and WhatsApp. As shown in Table~\ref{roust2}, our model consistently maintains significant performance advantages after online transmission. 

% Furthermore, as illustrated in Table~\ref{pf}, our model not only demonstrates outstanding performance but also features the fewest parameters and the lowest computational complexity.

\begin{table}[!t]
\centering
% \resizebox{0.47\textwidth}{!}{%
 \small % 这是唯一允许的字体大小调整命令
 \setlength{\tabcolsep}{1mm}
\begin{tabular}{cccccc}
\toprule \textbf{Method} & None & Facebook & WeiBo & WeChat & WhatsApp \\
\midrule
MFI-Net & 0.524 & 0.449 & 0.455 & 0.363 & 0.474 \\
SparesViT & 0.557 & 0.493 &  0.529 & 0.365 & 0.506 \\
PIM   & 0.548 &0.581 &0.566 &0.505 &0.585\\
Mesorch& 0.703  &  0.671  & 0.655  &   0.583  &  0.677  \\
 Ours &  \pmb{0.716} & \pmb{0.685} & \pmb{0.690}  &  \pmb{0.623}  &  \pmb{0.686}   \\
\bottomrule
\end{tabular}
% }
\caption{Robustness experiments on online social networks.}
\label{roust2}
\end{table}

% \begin{table}[!t]
% \centering
%  \small % 这是唯一允许的字体大小调整命令
%  \setlength{\tabcolsep}{1mm}
% % \resizebox{0.47\textwidth}{!}{%
% \begin{tabular}{cccccc}
% \toprule
% {\textbf{Metrics}} & {\textbf{MFI-Net}} & {\textbf{SparesViT}} & {\textbf{PIM}} & {\textbf{Mesorch}} & {\textbf{Ours}} \\ 

% \midrule
%  \textbf{Params (M)}& 32.54 & 50.30 & 152.47 & 85.75 & \pmb{27.57}\\

% \textbf{Flops (G)}& 36.25 & 46.20 & 682.88 & 124.93 & \pmb{35.39} \\ 

% \bottomrule
% \end{tabular}
% % }
% \caption{Comparison of model parameters and computational complexity.}
% \label{pf}
% \end{table}

\section{Conclusion}
In this work, we present and release the first scribble-annotated  IML dataset, Sc-IML, filling an important gap in weakly supervised annotation resources for the field. We also propose the first scribble-based weakly supervised IML framework, which incorporates structural consistency, prior-aware feature modulation, and gated adaptive fusion modules, significantly boosting model's robustness and localization accuracy. Moreover, the confidence-aware entropy minimization loss further enhances the model’s generalization in weakly supervised and unlabeled regions. Experimental results demonstrate that our approach consistently outperforms existing fully supervised methods on both in-distribution and out-of-distribution datasets. Our work provides a new perspective for low-cost annotation and weakly supervised learning in IML, and significantly advances the development of this field.

\section{Acknowledgments}
This work was supported in part by the National Natural Science Foundation of China under Grant 62302427, Grant 62462060, and Grant 62472368, in part by the Natural Science Foundation of Xinjiang Uygur Autonomous Region under Grant 2023D01C175.

\bibliography{aaai2026}

@article{wang2022pvt,
  title={Pvt v2: Improved baselines with pyramid vision transformer},
  author={Wang, Wenhai and Xie, Enze and Li, Xiang and Fan, Deng-Ping and Song, Kaitao and Liang, Ding and Lu, Tong and Luo, Ping and Shao, Ling},
  journal={Computational Visual Media},
  volume={8},
  number={3},
  pages={415--424},
  year={2022},
  publisher={Springer}
}

@article{liu2022pscc,
  title={PSCC-Net: Progressive spatio-channel correlation network for image manipulation detection and localization},
  author={Liu, Xiaohong and Liu, Yaojie and Chen, Jun and Liu, Xiaoming},
  journal={IEEE Transactions on Circuits and Systems for Video Technology},
  volume={32},
  number={11},
  pages={7505--7517},
  year={2022},
  publisher={IEEE}
}

@article{zhou2024contribution,
  title={A contribution-aware noise feature representation model for image manipulation localization},
  author={Zhou, Yang and Wang, Hongxia and Zeng, Qiang and Zhang, Rui and Meng, Sijiang},
  journal={Knowledge-Based Systems},
  pages={111988},
  year={2024},
  publisher={Elsevier}
}

@ARTICLE{ren2023mfi,
  author={Ren, Ruyong and Hao, Qixian and Niu, Shaozhang and Xiong, Keyang and Zhang, Jiwei and Wang, Maosen},
  journal={IEEE Transactions on Circuits and Systems for Video Technology}, 
  title={MFI-Net: Multi-Feature Fusion Identification Networks for Artificial Intelligence Manipulation}, 
  year={2024},
  volume={34},
  number={2},
  pages={1266-1280},
  doi={10.1109/TCSVT.2023.3289171}}

@inproceedings{dong2013casia,
  title={Casia image tampering detection evaluation database},
  author={Dong, Jing and Wang, Wei and Tan, Tieniu},
  booktitle={2013 IEEE China summit and international conference on signal and information processing},
  pages={422--426},
  year={2013},
  organization={IEEE}
}

@inproceedings{guan2019mfc,
  title={MFC datasets: Large-scale benchmark datasets for media forensic challenge evaluation},
  author={Guan, Haiying and Kozak, Mark and Robertson, Eric and Lee, Yooyoung and Yates, Amy N and Delgado, Andrew and Zhou, Daniel and Kheyrkhah, Timothee and Smith, Jeff and Fiscus, Jonathan},
  booktitle={2019 IEEE Winter Applications of Computer Vision Workshops (WACVW)},
  pages={63--72},
  year={2019},
  organization={IEEE}
}

@inproceedings{wen2016coverage,
  title={COVERAGE—A novel database for copy-move forgery detection},
  author={Wen, Bihan and Zhu, Ye and Subramanian, Ramanathan and Ng, Tian-Tsong and Shen, Xuanjing and Winkler, Stefan},
  booktitle={2016 IEEE international conference on image processing (ICIP)},
  pages={161--165},
  year={2016},
  organization={IEEE}
}

@article{hsu2006columbia,
  title={Columbia uncompressed image splicing detection evaluation dataset},
  author={Hsu, J and Chang, SF},
  journal={Columbia DVMM Research Lab},
  volume={6},
  year={2006}
}

@ARTICLE{10038722,
  author={Zhu, Jingru and Guo, Ya and Sun, Geng and Yang, Libo and Deng, Min and Chen, Jie},
  journal={IEEE Transactions on Geoscience and Remote Sensing}, 
  title={Unsupervised Domain Adaptation Semantic Segmentation of High-Resolution Remote Sensing Imagery With Invariant Domain-Level Prototype Memory}, 
  year={2023},
  volume={61},
  number={},
  pages={1-18},
  doi={10.1109/TGRS.2023.3243042}}

@inproceedings{guillaro2023trufor,
  title={Trufor: Leveraging all-round clues for trustworthy image forgery detection and localization},
  author={Guillaro, Fabrizio and Cozzolino, Davide and Sud, Avneesh and Dufour, Nicholas and Verdoliva, Luisa},
  booktitle={Proceedings of the IEEE/CVF conference on computer vision and pattern recognition},
  pages={20606--20615},
  year={2023}
}

@article{chen2024ean,
  title={EAN: Edge-Aware Network for Image Manipulation Localization},
  author={Chen, Yun and Cheng, Hang and Wang, Haichou and Liu, Ximeng and Chen, Fei and Li, Fengyong and Zhang, Xinpeng and Wang, Meiqing},
  journal={IEEE Transactions on Circuits and Systems for Video Technology},
  year={2024},
  publisher={IEEE}
}

@ARTICLE{10883001,
  author={Kong, Chenqi and Luo, Anwei and Wang, Shiqi and Li, Haoliang and Rocha, Anderson and Kot, Alex C.},
  journal={IEEE Transactions on Pattern Analysis and Machine Intelligence}, 
  title={Pixel-Inconsistency Modeling for Image Manipulation Localization}, 
  year={2025},
  volume={},
  number={},
  pages={1-18},
  doi={10.1109/TPAMI.2025.3541028}}

@inproceedings{zhu2025mesoscopic,
  title={Mesoscopic insights: orchestrating multi-scale \& hybrid architecture for image manipulation localization},
  author={Zhu, Xuekang and Ma, Xiaochen and Su, Lei and Jiang, Zhuohang and Du, Bo and Wang, Xiwen and Lei, Zeyu and Feng, Wentao and Pun, Chi-Man and Zhou, Ji-Zhe},
  booktitle={Proceedings of the AAAI Conference on Artificial Intelligence},
  volume={39},
  number={10},
  pages={11022--11030},
  year={2025}
}

@INPROCEEDINGS{10889843,
  author={Zhu, Zhangchen and Li, Jiafeng and Wen, Ying},
  booktitle={ICASSP 2025 - 2025 IEEE International Conference on Acoustics, Speech and Signal Processing (ICASSP)}, 
  title={Self-Optimization Training for Weakly Supervised Image Manipulation Localization}, 
  year={2025},
  volume={},
  number={},
  pages={1-5},
  doi={10.1109/ICASSP49660.2025.10889843}}

@article{ZHOU2024123501,
title = {Exploring weakly-supervised image manipulation localization with tampering Edge-based class activation map},
journal = {Expert Systems with Applications},
volume = {249},
pages = {123501},
year = {2024},
issn = {0957-4174},
doi = {https://doi.org/10.1016/j.eswa.2024.123501},
url = {https://www.sciencedirect.com/science/article/pii/S095741742400366X},
author = {Yang Zhou and Hongxia Wang and Qiang Zeng and Rui Zhang and Sijiang Meng}
}

@inproceedings{zhai2023towards,
  title={Towards Generic Image Manipulation Detection with Weakly-Supervised Self-Consistency Learning},
  author={Zhai, Yuanhao and Luan, Tianyu and Doermann, David and Yuan, Junsong},
  booktitle={Proceedings of the IEEE/CVF International Conference on Computer Vision},
  pages={22390--22400},
  year={2023}
}

@inproceedings{roth2022towards,
  title={Towards total recall in industrial anomaly detection},
  author={Roth, Karsten and Pemula, Latha and Zepeda, Joaquin and Sch{\"o}lkopf, Bernhard and Brox, Thomas and Gehler, Peter},
  booktitle={Proceedings of the IEEE/CVF conference on computer vision and pattern recognition},
  pages={14318--14328},
  year={2022}
}

@INPROCEEDINGS{Novozamsky_2020_WACV,
author = {Novozamsky, Adam and Mahdian, Babak and Saic, Stanislav},
title = {IMD2020: A Large-Scale Annotated Dataset Tailored for Detecting Manipulated Images},
booktitle = {2020 IEEE Winter Applications of Computer Vision Workshops (WACVW)},
year = {2020},
month = {March},
pages = {71-80}
}

@article{nichol2021glide,
  title={Glide: Towards photorealistic image generation and editing with text-guided diffusion models},
  author={Nichol, Alex and Dhariwal, Prafulla and Ramesh, Aditya and Shyam, Pranav and Mishkin, Pamela and McGrew, Bob and Sutskever, Ilya and Chen, Mark},
  journal={arXiv preprint arXiv:2112.10741},
  year={2021}
}

@inproceedings{korus2016evaluation,
  title={Evaluation of random field models in multi-modal unsupervised tampering localization},
  author={Korus, Pawe{\l} and Huang, Jiwu},
  booktitle={2016 IEEE international workshop on information forensics and security (WIFS)},
  pages={1--6},
  year={2016},
  organization={IEEE}
}

@inproceedings{huh2018fighting,
  title={Fighting fake news: Image splice detection via learned self-consistency},
  author={Huh, Minyoung and Liu, Andrew and Owens, Andrew and Efros, Alexei A},
  booktitle={Proceedings of the European conference on computer vision (ECCV)},
  pages={101--117},
  year={2018}
}

@article{bai2025weakly,
  title={Weakly-supervised cross-contrastive learning network for image manipulation detection and localization},
  author={Bai, Ruyi},
  journal={Knowledge-Based Systems},
  volume={310},
  pages={113033},
  year={2025},
  publisher={Elsevier}
}

@inproceedings{su2025can,
  title={Can we get rid of handcrafted feature extractors? sparsevit: Nonsemantics-centered, parameter-efficient image manipulation localization through spare-coding transformer},
  author={Su, Lei and Ma, Xiaochen and Zhu, Xuekang and Niu, Chaoqun and Lei, Zeyu and Zhou, Ji-Zhe},
  booktitle={Proceedings of the AAAI Conference on Artificial Intelligence},
  volume={39},
  number={7},
  pages={7024--7032},
  year={2025}
}

@article{kong2025pixel,
  title={Pixel-inconsistency modeling for image manipulation localization},
  author={Kong, Chenqi and Luo, Anwei and Wang, Shiqi and Li, Haoliang and Rocha, Anderson and Kot, Alex C},
  journal={IEEE Transactions on Pattern Analysis and Machine Intelligence},
  year={2025},
  publisher={IEEE}
}

@article{obukhov2019gated,
  title={Gated CRF loss for weakly supervised semantic image segmentation},
  author={Obukhov, Anton and Georgoulis, Stamatios and Dai, Dengxin and Van Gool, Luc},
  journal={arXiv preprint arXiv:1906.04651},
  year={2019}
}

@inproceedings{he2023weakly,
  title={Weakly-supervised camouflaged object detection with scribble annotations},
  author={He, Ruozhen and Dong, Qihua and Lin, Jiaying and Lau, Rynson WH},
  booktitle={Proceedings of the AAAI conference on artificial intelligence},
  volume={37},
  number={1},
  pages={781--789},
  year={2023}
}

@ARTICLE{MVSS_2022TPAMI,
  author={Dong, Chengbo and Chen, Xinru and Hu, Ruohan and Cao, Juan and Li, Xirong},
  journal={IEEE Transactions on Pattern Analysis and Machine Intelligence}, 
  title={MVSS-Net: Multi-View Multi-Scale Supervised Networks for Image Manipulation Detection}, 
  year={2022},
  volume={},
  number={},
  pages={1-14},
  doi={10.1109/TPAMI.2022.3180556}
}

@article{ma2025imdl,
  title={Imdl-benco: A comprehensive benchmark and codebase for image manipulation detection \& localization},
  author={Ma, Xiaochen and Zhu, Xuekang and Su, Lei and Du, Bo and Jiang, Zhuohang and Tong, Bingkui and Lei, Zeyu and Yang, Xinyu and Pun, Chi-Man and Lv, Jiancheng and others},
  journal={Advances in Neural Information Processing Systems},
  volume={37},
  pages={134591--134613},
  year={2025}
}

@article{10.1504/ijaacs.2024.139383,
author = {Gu, Fei and Dai, Yunshu and Fei, Jianwei and Chen, Xianyi},
title = {Deepfake detection and localisation based on illumination inconsistency},
year = {2024},
issue_date = {2024},
publisher = {Inderscience Publishers},
address = {Geneva 15, CHE},
volume = {17},
number = {4},
issn = {1754-8632},
url = {https://doi.org/10.1504/ijaacs.2024.139383},
doi = {10.1504/ijaacs.2024.139383},
journal = {Int. J. Auton. Adapt. Commun. Syst.},
month = jan,
pages = {352–368},
numpages = {16}
}

@article{10.1504/ijaacs.2024.142523,
author = {He, Guangyang and Zhang, Xiang and Wang, Fan and Fu, Zhangjie},
title = {A novel copy-move detection and location technique based on tamper detection and similarity feature fusion},
year = {2024},
issue_date = {2024},
publisher = {Inderscience Publishers},
address = {Geneva 15, CHE},
volume = {17},
number = {6},
issn = {1754-8632},
url = {https://doi.org/10.1504/ijaacs.2024.142523},
doi = {10.1504/ijaacs.2024.142523},
journal = {Int. J. Auton. Adapt. Commun. Syst.},
month = jan,
pages = {514–529},
numpages = {15}
}

\end{document}